\documentclass[conference]{IEEEtran}
\IEEEoverridecommandlockouts           

\usepackage{amsmath,amssymb}
\usepackage{graphicx}
\usepackage{booktabs}
\usepackage{array}
\usepackage{cite}
\usepackage{tikz}
\usepackage{xcolor}
\usepackage{balance}
\usepackage{caption}
\usepackage{booktabs}
\usepackage{tabularx}
\usepackage{array}
\usetikzlibrary{arrows.meta,positioning,fit,calc}
\usepackage{booktabs}
\usepackage{tabularx}
\usepackage{array}
\usepackage{makecell}

\newcommand{\TopK}{\operatorname{TopK}}
\newcommand{\ADE}{\operatorname{ADE}}
\newcommand{\FDE}{\operatorname{FDE}}

\title{Mind the Privileged-to-Camera Gap: Actor-Centric Sidecar Supervision for Camera-First Open-Loop Waypoint Prediction}

\author{Feeza Khan Khanzada and Jaerock Kwon

\thanks{Feeza Khan Khanzada and Jaerock Kwon are with the Department of Electrical and Computer Engineering, University of Michigan-Dearborn, 4901 Evergreen Rd, Dearborn, MI 48128, United States. {\tt\small \{feezakk, jrkwon\}@umich.edu}}%
}

\begin{document}
\maketitle

\begin{abstract}
Camera-first autonomous-driving models predict future ego waypoints from images, ego-state features, and route commands, but waypoint supervision alone does not explicitly supervise actor-level representations of nearby road users. We study this as supervised representation learning for open-loop waypoint prediction. The deployable model uses multi-view RGB, ego state, and route command at inference. During training, simulator-derived sidecar labels supervise actor grounding, privileged hindsight actor relevance relative to the logged ego trajectory, and selected-actor short-horizon motion; these labels are never inference inputs. We evaluate route-disjoint splits with matched architecture, optimizer, validation criterion, checkpoint selection, and three seeds. A plain waypoint-only RGB baseline obtains 1.815$\pm$0.02 m final displacement error (FDE), and the matched no-teacher non-sidecar RGB control obtains 1.716$\pm$0.02 m. Road-user sidecar supervision (RU-sidecar) reduces FDE to 1.223$\pm$0.01 m, a 32.6\% reduction over the plain baseline and 28.7\% over the matched no-teacher non-sidecar RGB control. It improves over the plain baseline on 1445/1494 routes and over the matched no-teacher non-sidecar RGB control on 1417/1494 routes. Actor-conditioned slices show gains in all nonempty subsets, including 29.1\% reduction for samples with at least four valid sidecar actors and 30.0\% when a vulnerable road user is present. Optional simulator-state teacher alignment reaches 1.186$\pm$0.15 m FDE, but higher seed variability makes it secondary. Non-deployable simulator-state diagnostics remain stronger, indicating a privileged-to-camera gap. The evidence is limited to open-loop simulation diagnostics.
\end{abstract}

\section{Introduction}

Camera-first driving models are attractive because they use inputs available on real vehicles, including camera images, ego-state measurements, and high-level route commands. Many end-to-end systems map these inputs directly to future ego waypoints or low-level controls~\cite{pmlr-v100-chen20a,10.5555/3600270.3600713}. This direct formulation is simple and scalable, but it can leave the structure of nearby road users weakly supervised. In urban scenes, the ego trajectory often depends on a small set of vehicles, pedestrians, or cyclists. A waypoint-only objective constrains the final path but may not force intermediate image features to encode which actors exist, which ones are relevant, or how they move. We therefore treat camera-first waypoint prediction as a supervised actor-centric representation-learning problem. Our approach adds simulator-derived sidecar labels during training to supervise actor grounding, target-conditioned actor relevance, and selected-actor short-horizon motion. These labels are auxiliary training targets only; at inference, the deployable model still uses only RGB observations, ego-state features, and a route command. 

We test whether road-user sidecar supervision (RU-sidecar) improves RGB-only open-loop waypoint prediction under a controlled protocol. We use the term student interface to denote the actor-centric RGB architecture in Fig. \ref{fig:methodology}, which predicts actor slots, ranks actor relevance, predicts selected-actor futures, and uses those actor-future tokens for waypoint prediction. All student-interface RGB configurations use this same architecture and inference inputs. When used, teacher alignment denotes training-time distillation from a non-deployable simulator-state model; teacher outputs are never inference inputs. The key controlled comparison is a waypoint-student/no-teacher configuration, which keeps the same student interface but disables both sidecar losses and teacher alignment. We also report a conventional waypoint-only RGB baseline that does not use the actor-slot interface, and we include non-deployable simulator-state diagnostics to estimate the value of exact road-user state.

On held-out routes, the plain waypoint-only RGB baseline obtains 1.815 $\pm$ 0.02 m test final displacement error (FDE), and the matched no-teacher non-sidecar RGB control obtains 1.716 $\pm$ 0.02 m. The RU-sidecar model without teacher alignment reduces test FDE to 1.223 $\pm$ 0.01 m, corresponding to a 32.6\% reduction relative to the plain baseline and a 28.7\% reduction relative to the matched no-teacher non-sidecar RGB control. It improves over the plain baseline on 1445/1494 test routes and over the matched no-teacher non-sidecar RGB control on 1417/1494 routes. Optional simulator-state teacher alignment obtains 1.186 $\pm$ 0.15 m FDE, but its larger seed variability and small mean gain over the no-teacher model make it secondary. Actor-conditioned slice analysis shows improvements in all nonempty actor-conditioned slices, including actor-dense scenes and scenes containing a vulnerable road user (VRU), such as a pedestrian or cyclist. Because evaluation is open-loop, our claim is limited to offline waypoint prediction.

The main contributions are:
\begin{itemize}
\item We formulate camera-first open-loop waypoint prediction as an actor-centric representation-learning problem, where RGB features are trained to expose road-user structure before ego waypoint prediction.

\item We introduce simulator-derived, training-only RU-sidecar labels for actor grounding, hindsight actor relevance, and selected-actor short-horizon motion while preserving RGB-only inference.

\item We evaluate a matched control ladder that separates conventional waypoint imitation, waypoint-loss reweighting, shuffled-teacher control, valid teacher alignment, and RU-sidecar supervision under the same route-disjoint protocol.

\item We show that RU-sidecar without teacher alignment reduces FDE from 1.815±0.02 m to 1.223±0.01 m and improves over the matched no-teacher non-sidecar control on 1417/1494 held-out routes, while non-deployable diagnostics quantify the remaining privileged-to-camera gap.
\end{itemize}

\section{Related Work}

End-to-end driving is commonly formulated as imitation learning from camera observations, route commands, ego-state features, or fused sensor inputs to controls or future waypoints. Conditional imitation learning introduced command-conditioned visuomotor policies for route-aware driving~\cite{10.1109/ICRA.2018.8460487}. Subsequent CARLA-based systems improved closed-loop performance through privileged supervision, reinforcement-learning teachers, multimodal fusion, trajectory-control coupling, and stronger representation learning~\cite{pmlr-v100-chen20a,9711506,9863660,10.5555/3600270.3600713,chen2022lav}. However, these studies generally do not isolate, under a matched camera-first architecture, whether training-only actor-level supervision improves RGB-only waypoint prediction or how large the remaining gap is to privileged actor-state inputs. Our work studies this question in an open-loop setting. We evaluate whether RU-sidecar supervision for actor grounding,
target-conditioned actor relevance, and selected-actor motion improves
RGB-only waypoint prediction, and we use non-deployable actor-state diagnostics to estimate the remaining gap to privileged road-user state.

A related line of work uses object-level or actor-level structure for planning instead of relying only on dense image, BEV, or latent scene features. Object-centric planners such as PlanT show that planning can be expressed over actor tokens rather than dense grids~\cite{Renz2022CORL}. Actor-level supervision also appears in reinforcement-learning, coach and all-vehicle supervision frameworks, where surrounding agents provide additional structure for learning driving behavior~\cite{9711506,chen2022lav}. These works motivate actor-level structure, but they often use actor tokens, actor state, or full-scene agent supervision as part of the policy-learning setup, and they are usually evaluated as complete driving systems. In contrast, our deployable model predicts actor-token representations from RGB and uses simulator actor state only offline to construct training labels. We isolate an open-loop question by testing whether training-only sidecars for actor grounding, target-conditioned actor relevance, and selected-actor motion improve RGB-only ego-waypoint prediction.

Privileged-supervision and world-model methods are complementary: they use privileged policies, actor state, or predictive latent scene objectives to improve full driving systems. Learning by Cheating transfers from a privileged policy to a vision policy, LAV supervises behavior using all vehicles, and world-model methods learn predictive latent scene representations \cite{pmlr-v100-chen20a,chen2022lav,mile2022}. In contrast, we isolate a narrower representation-learning question: under a fixed RGB-only waypoint-prediction interface, does RU-sidecar improve open-loop ego-waypoint prediction, and how large is the remaining gap to exact simulator actor state? This framing separates our contribution from privileged-policy transfer, actor-token planning systems, and latent world-model training.
\section{Method}

Our goal is to train a camera-first waypoint predictor that can use actor-level supervision during training without requiring actor state at inference. We use actor to denote a nearby road user, such as a vehicle, pedestrian, cyclist, or other traffic participant. We use sidecar label to denote an auxiliary supervised target generated offline from simulator state and used only during training or diagnostic evaluation. The method separates three components. The deployable camera path predicts actor-centric intermediate representations from RGB, ego-state features, and a route command. The offline label generator converts simulator state into actor grounding, actor relevance, and actor-future sidecar labels. The non-deployable simulator-state path encodes exact simulator actor state and is used only as a diagnostic reference and, when enabled, as an optional teacher for camera-based models.

\subsection{Overview}
\begin{figure*}
    \captionsetup{font=footnotesize,skip=2pt}
    \centering
    \includegraphics[width=1.0\textwidth]{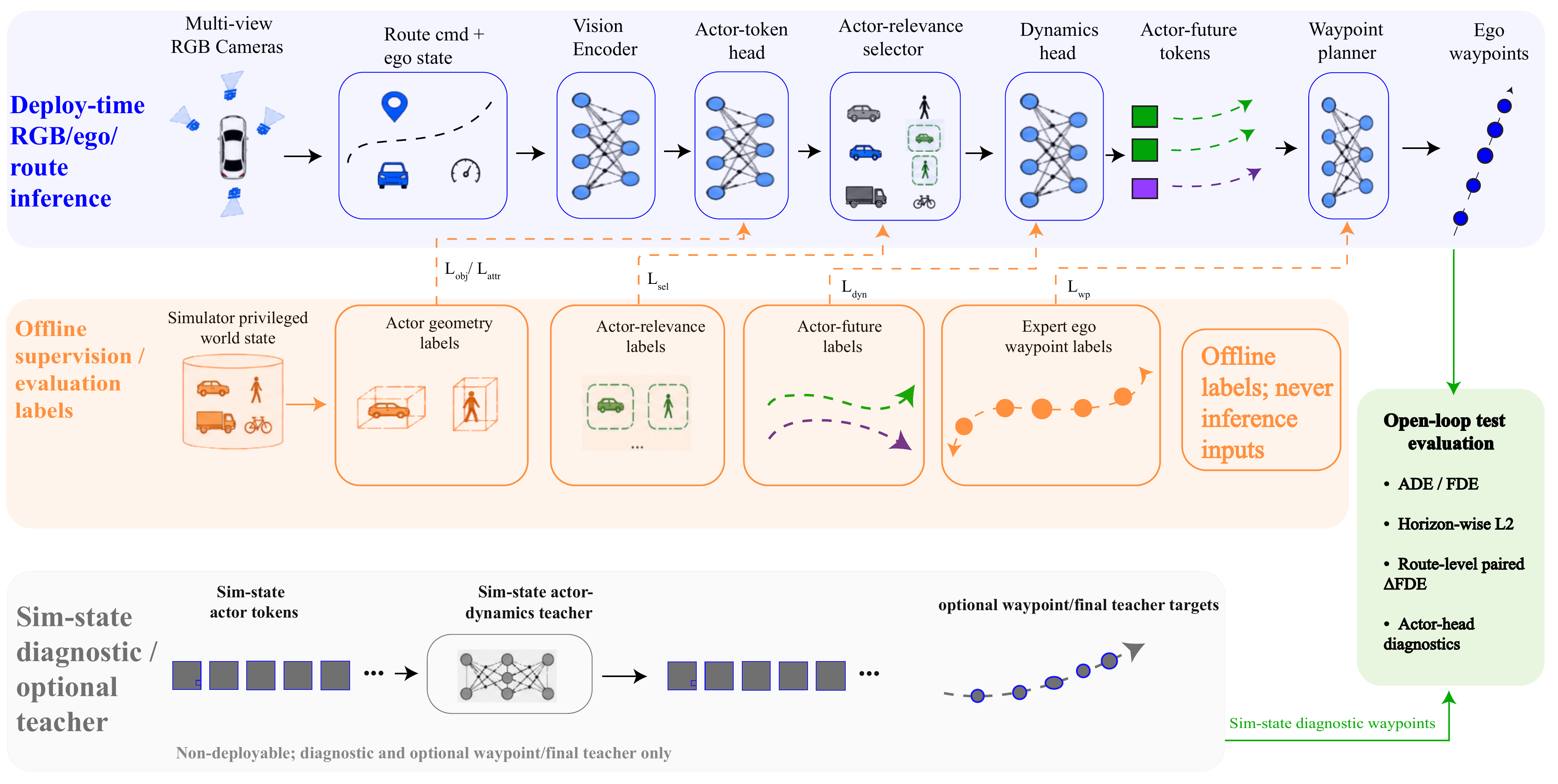}
    \caption{Camera-first road-user-supervised open-loop waypoint prediction. At inference, every deployable model uses only multi-view RGB observations, ego state, and route command. Simulator-derived sidecars provide actor geometry, image-grounded actor labels, target-conditioned actor-relevance labels, and selected-actor future labels for training and diagnostics. All student-interface camera-based configurations share the same architecture; configurations differ by active training signals. The plain waypoint-only RGB baseline is included as a conventional waypoint-imitation reference. The simulator-state pathway is non-deployable and is used only for diagnostics and optional teacher alignment.}
    \label{fig:methodology}
\end{figure*}

Figure~\ref{fig:methodology} summarizes the resulting framework. At inference, every deployable camera-based configuration receives only multi-view RGB observations, ego-state features, and a route command. The RGB history is encoded into visual features, from which an actor-token head predicts a fixed-size set of camera-derived actor tokens. A selector ranks these tokens by relevance to the ego vehicle, and a dynamics head predicts short-horizon futures for the selected actors. The waypoint planner attends to the resulting actor-future tokens and predicts future ego waypoints.

The bottom branch of Figure~\ref{fig:methodology} shows the non-deployable simulator-state reference. It uses exact simulator actor state and is included only for diagnostics and optional teacher alignment; it is not a camera-first deployable model.

\subsection{Task Formulation} \label{sec:task_formulation}

At time \(t\), a deployable camera-first model observes a history of multi-view RGB images,
\begin{equation}
I_{t-H+1:t}=\{I_{\tau}^{v}: \tau=t-H+1,\ldots,t,\; v\in\mathcal{V}\}
\end{equation}
where \(\mathcal{V}\) is the camera-view set and \(H\) is the history length. The model also receives ego-state features \(e_t\) and a route command \(r_t\). Its inference input is therefore
\begin{equation}
x_t = (I_{t-H+1:t}, e_t, r_t)
\end{equation}

The primary supervised target is a sequence of \(T=6\) future ego waypoints in the ego frame at time \(t\),
\begin{equation}
Y_t=\{y_{t+\Delta}\}_{\Delta=1}^{T},\qquad y_{t+\Delta}\in\mathbb{R}^2
\end{equation}
and the model predicts
\begin{equation}
\hat{Y}_t=f_\theta(x_t)=\{\hat{y}_{t+\Delta}\}_{\Delta=1}^{T}
\end{equation}

During training, each sample also has an offline simulator-derived supervision record. This record contains road-user states
\begin{equation}
A_t=\{a_{i,t}\}_{i=1}^{N_t}
\end{equation}
where \(a_{i,t}\) includes class, ego-relative geometry, heading, velocity or motion proxy, distance, and validity masks when available. These actor states are not inference inputs for deployable models. They are used offline to construct auxiliary sidecar labels for actor grounding, target-conditioned actor relevance, and selected-actor short-horizon motion. We denote the auxiliary sidecar labels by

\begin{equation}
S_t = (B_t, G_t, C_t, F_t)
\end{equation}

where \(B_t\) contains actor-geometry and attribute labels, \(G_t\) contains image-grounded actor labels, \(C_t\) contains target-conditioned actor-relevance labels, and \(F_t\) contains future-motion targets for selected actors. The deployable training objective uses \(Y_t\) as the primary waypoint-imitation target and \(S_t\) as auxiliary sidecar supervision. At evaluation time, deployable models receive only \(x_t\), not \(A_t\), \(S_t\), future actor states, or future ego waypoints.

\subsection{Road-User Sidecar Label Construction}

Sidecar labels are generated offline from simulator state and are used only as auxiliary training targets. The generator outputs four groups. First, actor-geometry and attribute labels \(B_t\) describe each valid simulator road user by class, ego-frame position, bounding geometry, heading, velocity or motion proxy, distance, and validity mask. Second, image-grounded labels \(G_t\) are produced by projecting simulator actors into the multi-view camera rig. For projected actors, the label stores the camera view, image-space box or center, class, depth, ego-frame position, and validity mask. Third, target-conditioned relevance labels \(C_t\) rank valid actors by their relevance to the logged ego behavior. The relevance score combines distance, relative motion, time-to-conflict, route-lane relevance, vulnerable-road-user class, and future ego-path overlap:
{\small
\[
c_i = w_d c_i^{dist} + w_v c_i^{relvel} + w_{ttc} c_i^{ttc}
+ w_\ell c_i^{lane} + w_{vru} c_i^{vru} + w_o c_i^{overlap}
\]
}




Each component is normalized to \([0,1]\). Actors are ranked by \(c_i\) after validity filtering, with ties broken by ego-relative distance. The top \(K\) valid actors define the selected actor set. The future ego-path overlap component uses the logged future ego trajectory and is therefore a hindsight supervision label, not an input feature. It is computed offline only to construct the actor-relevance target. At validation and test inference, deployable models receive neither future ego waypoints, future actor states, nor simulator actor tables; they must infer relevance from RGB, ego state, and route command. This target-conditioned label asks whether camera features can learn to represent actors associated with the demonstrated ego trajectory without changing the deployable input interface.

Fourth, actor-future labels \(F_t\) store short-horizon future states for selected actors,
\[
F_{i,t} = \{f_{i,t+\Delta}\}_{\Delta=1}^{T_a},
\]
with masks for actors that disappear or lack reliable annotation. Expert ego waypoints provide the primary waypoint-imitation target and the open-loop evaluation target. Section \ref{sec:implementation_details} reports the fixed weights, thresholds, and slot-matching rule used by the label generator.

\subsection{Camera-First Actor-Centric Model}

The deployable model predicts ego waypoints from RGB history, ego-state features, and a route command. A visual encoder maps the multi-view RGB history to visual features,
\begin{equation}
V_t=\psi_{\mathrm{img}}(I_{t-H+1:t})
\end{equation}
In our implementation, \(\psi_{\mathrm{img}}\) is the shared multi-view RGB encoder described in Section~\ref{sec:implementation_details}. It produces visual features that are used by the actor-token branch and waypoint planner.

The actor-token head predicts a fixed-size set of camera-derived actor-slot tokens and associated actor attributes,
\begin{equation}
(O_t^{\mathrm{cam}}, P_t^{\mathrm{cam}}, \hat{m}_t^{\mathrm{cam}})
=\psi_{\mathrm{obj}}(V_t,e_t,r_t)
\end{equation}
where \(O_t^{\mathrm{cam}}\in\mathbb{R}^{N\times d}\) are actor-slot tokens, \(P_t^{\mathrm{cam}}\) contains predicted actor attributes, and \(\hat{m}_t^{\mathrm{cam}}\) contains objectness or slot-validity logits. The predicted attributes include an ego-frame center \(\hat{\mathbf p}^{xy}_{i,t}\) for each actor slot.

During training, actor-sidecar losses are applied after matching predicted actor slots to sidecar actors. We use deterministic greedy one-to-one nearest-center matching in ego-frame coordinates with cost
\[
c_{ij}=\|\hat{\mathbf p}^{xy}_{i,t}-\mathbf p^{xy}_{j,t}\|_2 
\]
where \(\hat{\mathbf p}^{xy}_{i,t}\) and \(\mathbf p^{xy}_{j,t}\) are the predicted and sidecar ego-frame actor centers. Valid slot/actor pairs are sorted by \((c_{ij},i,j)\) and accepted greedily when both the slot and actor are unused. Unmatched predicted slots receive negative objectness targets and are excluded from attribute, selector, and dynamics regression losses. Unmatched sidecar actors do not contribute regression losses for that sample. 


For actor tokens \(O_t=\{o_{i,t}\}_{i=1}^{N}\), the selector predicts sidecar-defined actor-relevance scores,
\begin{equation}
\hat{s}_{i,t}=g_{\theta}(o_{i,t},e_t,r_t),
\qquad
\mathcal{K}_t=\TopK(\hat{s}_{1,t},\ldots,\hat{s}_{N,t})
\end{equation}
The selected set \(\mathcal{K}_t\) contains the \(K\) actor slots used for actor-future prediction and waypoint planning. The waypoint planner consumes the model-selected top-\(K\) actor slots during both training and inference; sidecar labels are used only to supervise the selector and dynamics losses.

The dynamics head predicts short-horizon futures for the selected actors,
\begin{equation}
\hat{F}_{t,K}=h_\theta(O_{t,K}, e_t, r_t) \in \mathbb{R}^{K \times T_a \times d_a}
\end{equation}

where $d_a$ denotes the actor-future output dimension, including ego-frame actor position, heading, and velocity or motion-proxy components used by the dynamics loss. These predicted futures are embedded together with the selected actor tokens,
\begin{equation}
Z_t^{\mathrm{dyn}}=\eta_{\theta}(O_{t,\mathcal{K}},\hat{F}_{t,\mathcal{K}})
\end{equation}

The waypoint planner constructs an ego query \(q_t^{\mathrm{ego}}=\rho_{\theta}(e_t,r_t)\), attends to the actor-future tokens, and predicts \(T\) future ego waypoints:
\begin{align}
 c_t^{\mathrm{ego}} &=
 \operatorname{CrossAttn}
 \left(q_t^{\mathrm{ego}},
 \operatorname{Key}(Z_t^{\mathrm{dyn}}),
 \operatorname{Value}(Z_t^{\mathrm{dyn}})\right) \\
 \hat{Y}_t &= d_{\theta}([q_t^{\mathrm{ego}},c_t^{\mathrm{ego}}])
\end{align}

\subsection{Simulator-State Diagnostic and Optional Teacher}

The simulator-state pathway is a non-deployable reference model that replaces RGB-based actor recovery with exact simulator actor state available at time \(t\). It encodes the actor table \(A_t\) into actor tokens,
\begin{equation}
O_t^{\mathrm{sim}}=\phi_{\mathrm{obj}}(A_t)
\end{equation}
These tokens are passed to the same actor-conditioning and waypoint-prediction interface used by the camera-first model. This pathway is not a camera-first deployable system because \(A_t\) is an exact simulator actor table rather than information recovered from RGB.

We use this pathway for two purposes. First, it provides a diagnostic reference for estimating how much open-loop waypoint prediction improves when current road-user state is already known accurately. Second, when teacher alignment is enabled, a fixed simulator-state teacher provides waypoint and final-horizon targets for the camera-based student. These teacher predictions are used only as training targets for teacher-aligned configuration and are never available to deployable models at inference.

We also report an oracle sidecar-selected actor diagnostic$\dagger$. Unlike deployable RU-sidecar models, this diagnostic does not
predict actor relevance from RGB. It receives the sidecar-selected actor set directly and passes those actor tokens to the same actor-conditioned waypoint interface. This diagnostic estimates
the value of sidecar-defined actor selection independent of camera-side actor recovery; it is not a deployable camera-first
model.

\subsection{Training Objectives}

Training uses the primary waypoint-imitation target together with optional actor-centric sidecar losses. The total objective is
\begin{equation}
\begin{aligned}
\mathcal{L}
={}&
\lambda_{\mathrm{wp}}\mathcal{L}_{\mathrm{wp}}
+\lambda_{\mathrm{obj}}\mathcal{L}_{\mathrm{obj}}
+\lambda_{\mathrm{attr}}\mathcal{L}_{\mathrm{attr}} \\
&+
\lambda_{\mathrm{sel}}\mathcal{L}_{\mathrm{sel}}
+\lambda_{\mathrm{dyn}}\mathcal{L}_{\mathrm{dyn}}
+\lambda_{\mathrm{fwd}}\mathcal{L}_{\mathrm{fwd}}
+\mathcal{L}_{\mathrm{dist}} 
\end{aligned}
\end{equation}

The waypoint loss is the primary supervised objective,
\begin{equation}
\mathcal{L}_{\mathrm{wp}}
=
\frac{1}{|\Omega_Y|}
\sum_{(t,\Delta)\in\Omega_Y}
\ell(\hat{y}_{t+\Delta},y_{t+\Delta})
\end{equation}
where \(\ell\) is the SmoothL1 regression loss in ego-frame meters and \(\Omega_Y\) contains valid waypoint targets.

The sidecar losses supervise intermediate actor structure. \(\mathcal{L}_{\mathrm{obj}}\) supervises actor-slot objectness, class, and image-space geometry. \(\mathcal{L}_{\mathrm{attr}}\) supervises continuous actor attributes such as depth and ego-frame position. \(\mathcal{L}_{\mathrm{sel}}\) supervises target-conditioned actor relevance, including ranking supervision when enabled. \(\mathcal{L}_{\mathrm{dyn}}\) supervises selected-actor short-horizon motion:
\begin{equation}
\mathcal{L}_{\mathrm{dyn}}
=
\frac{1}{|\Omega_F|}
\sum_{(i,t,\Delta)\in\Omega_F}
\left\|
\hat{f}_{i,t+\Delta}-f_{i,t+\Delta}
\right\|_1 
\end{equation}
where \(\Omega_F\) contains valid matched actor-future targets. \(\mathcal{L}_{\mathrm{fwd}}\) is a waypoint auxiliary that emphasizes later prediction horizons.

When a simulator-state teacher is used, the optional distillation term is
\begin{equation}
\begin{aligned}
\mathcal{L}_{\mathrm{dist}}
={}&
\lambda_{\mathrm{lat}}\mathcal{L}_{\mathrm{lat}}
+
\lambda_{\mathrm{wpd}}\mathcal{L}_{\mathrm{wpd}}
+
\lambda_{\mathrm{final}}\mathcal{L}_{\mathrm{final}} \\
&+
\lambda_{\mathrm{fwdT}}\mathcal{L}_{\mathrm{fwdT}} 
\end{aligned}
\end{equation}
In the reported teacher-aligned runs, only waypoint and final-horizon teacher targets are active; latent and forward-teacher alignment terms are implemented but assigned zero weight. No-teacher variants set \(\mathcal{L}_{\mathrm{dist}}=0\).

All student-interface RGB configurations use the same forward architecture and inference inputs. Within this student-interface family, configurations differ by active loss weights, teacher targets, or sidecar supervision, not by removing model modules. The conventional waypoint-only RGB baseline is reported separately. The exact loss weights and targeted ablation settings are reported in Section~\ref{sec:evaluated_configurations}.



\section{Experimental Setup}

\subsection{Dataset and Evaluation Protocol}
\label{sec:dataset_and_splits}

All experiments use a fixed simulator-based driving snapshot built from LMDrive~\cite{shao2023lmdrive}. The snapshot contains 4,216,721 sample records and 12,405 routes collected across multiple simulator towns. Routes are split at the route level so that no route appears in more than one of the training, validation, or test splits. After validity filtering, evaluation contains 465,095 validation samples and 467,728 test samples, with 1,494 routes in each held-out split. The training split contains \(9{,}417\) routes and \(3{,}283{,}898\) valid samples.

Each sample provides the deployable input \(x_t\), waypoint target \(Y_t\), and sidecar record \(S_t\) defined in Section~\ref{sec:task_formulation}. Deployable models receive only \(x_t\) at inference. The split is route-disjoint, not town-disjoint, weather-disjoint, simulator-disjoint, or real-world-disjoint. The evaluation therefore measures held-out-route generalization within one fixed simulator snapshot, not cross-domain or real-world transfer.

\subsection{Configurations and Loss Weights}
\label{sec:evaluated_configurations}

All student-interface deployable RGB configurations share the same architecture, trainable parameter count, optimizer, validation criterion, checkpoint-selection rule, and route-disjoint splits. They differ only in active loss terms, teacher-target pairing, or exported diagnostic outputs. The plain waypoint-only RGB baseline is a conventional waypoint-imitation reference; the waypoint-student/no-teacher configuration is the matched non-sidecar
control for the RU-sidecar models.

We evaluate seven deployable RGB configurations: waypoint-only RGB, forward/final-horizon weighted RGB, shuffled-teacher waypoint control, waypoint student without teacher alignment, waypoint student with teacher alignment, RU-sidecar without teacher alignment, and RU-sidecar with teacher alignment. We also evaluate two targeted no-teacher ablations of the road-user-supervised model: no actor-future loss and no selector loss.

The full road-user recipe uses
\(\lambda_{\mathrm{wp}}=1.0\),
\(\lambda_{\mathrm{obj}}=1.0\),
\(\lambda_{\mathrm{attr}}=0.5\),
\(\lambda_{\mathrm{sel}}=1.0\),
\(\lambda_{\mathrm{dyn}}=1.0\), and
\(\lambda_{\mathrm{fwd}}=0.5\).
The forward/final-horizon baseline activates only \(\lambda_{\mathrm{fwd}}=0.5\) in addition to waypoint imitation. The no-actor-future and no-selector ablations set \(\lambda_{\mathrm{dyn}}=0\) and \(\lambda_{\mathrm{sel}}=0\), respectively, while keeping the rest of the road-user recipe fixed. Teacher-aligned rows use \(\lambda_{\mathrm{wpd}}=\lambda_{\mathrm{final}}=0.05\) and \(\lambda_{\mathrm{lat}}=\lambda_{\mathrm{fwdT}}=0\); no-teacher rows set \(\mathcal{L}_{\mathrm{dist}}=0\).

The two non-deployable diagnostics are the oracle sidecar-selected actor diagnostic$^\dagger$ and the simulator-state actor-dynamics diagnostic$^\dagger$. Both receive simulator-derived actor information at evaluation and are therefore not valid camera-first deployable systems. They are included only to estimate how much waypoint prediction benefits from reliable actor information and how much privileged-to-camera gap remains. The control ladder separates waypoint-loss reweighting, invalid teacher association, valid teacher alignment, and RU-sidecar. It is not a full factorial ablation, so claims about road-user supervision refer to the complete sidecar recipe.

The waypoint-only RGB configuration is the plain waypoint-imitation baseline. The waypoint-student configurations use the same student interface as the teacher-aligned and RU-sidecar configurations but disable road-user sidecar losses; the no-teacher waypoint student also disables distillation. The
waypoint-student/no-teacher configuration is the matched no-teacher non-sidecar deployable RGB control. We report improvements relative to both the plain waypoint-only baseline and this matched control to avoid attributing gains to architectural or training-interface differences.

\begin{table}[t]
\centering
\captionsetup{font=footnotesize,skip=2pt}
\caption{Definition of deployable RGB controls. Student interface denotes the actor-slot, selector, actor-future, and actor-conditioned waypoint-planner interface in Fig. \ref{fig:methodology}. All deployable rows use only RGB, ego state, and route command at inference.}
\label{tab:deployable_rgb_controls}

\scriptsize
\setlength{\tabcolsep}{2pt}
\renewcommand{\arraystretch}{1.08}
\setlength{\arrayrulewidth}{0.4pt}

\begin{tabularx}{\columnwidth}{|>{\raggedright\arraybackslash}X|c|c|c|}
\hline
Configuration
& \makecell[c]{Student interface}
& \makecell[c]{Sidecar losses}
& \makecell[c]{Teacher loss} \\
\hline

Plain waypoint-only RGB & No & No & No \\
\hline
Forward/final weighted RGB & No & No & No \\
\hline
Shuffled-teacher control & Yes & No & Shuffled \\
\hline
Waypoint student, no teacher & Yes & No & No \\
\hline
Waypoint student, teacher-aligned & Yes & No & Yes \\
\hline
RU-sidecar, no teacher & Yes & Yes & No \\
\hline
RU-sidecar, teacher-aligned & Yes & Yes & Yes \\
\hline
\end{tabularx}
\end{table}

\subsection{Training, Checkpointing, and Metrics}

Each deployable RGB configuration is trained with three random seeds controlling initialization, data order, and stochastic training operations. For each seed and configuration, the best checkpoint is selected by validation FDE only, and the test set is evaluated once using that checkpoint. We report mean \(\pm\) sample standard deviation over the three seeds.

The primary open-loop waypoint metrics are average displacement error and final displacement error:
\begin{align}
\ADE &= \frac{1}{|\Omega|}\sum_{t\in\Omega}\frac{1}{T}\sum_{\Delta=1}^{T}\|\hat{y}_{t+\Delta}-y_{t+\Delta}\|_2 \\
\FDE &= \frac{1}{|\Omega|}\sum_{t\in\Omega}\|\hat{y}_{t+T}-y_{t+T}\|_2
\end{align}
where \(\Omega\) is the valid evaluated sample set. We also report horizon-wise waypoint L2 error.

For route-level analysis, each model is first compared against the same-seed plain waypoint-only RGB baseline on each route. Positive route-level $\Delta$FDE means the model has lower route-averaged FDE than the baseline on that route. Route win counts are computed after averaging paired route improvements across the three seeds, and confidence intervals are obtained by bootstrap resampling routes. We also report a stricter paired comparison in which the RU-sidecar model is compared against the matched no-teacher non-sidecar RGB control. Actor-selection recall, actor-selection precision, and actor-future ADE/FDE are reported only as auxiliary diagnostics.

\subsection{Implementation Details}
\label{sec:implementation_details}

Student-interface deployable RGB configurations use the same medium-size camera-first architecture. The conventional waypoint-only and forward/final weighted RGB baselines are reported separately because they do not instantiate the actor-slot, selector, actor-future, and actor-conditioned waypoint-planner interface. The student-interface architecture uses ConvNeXt-Tiny as the image encoder and a Transformer waypoint planner with hidden dimension 512, 8 layers, 8 attention heads, MLP ratio 4, and dropout 0.1. Each RGB view is resized to \(224\times224\). The input uses four camera views: front, left, right, and rear. The history length is \(H=3\), and the ego waypoint horizon is \(T=6\).

The shared student-interface architecture contains \(N=16\) camera actor slots, selects \(K=5\) actors, and uses actor-future horizon \(T_a=6\). All trainable configurations use AdamW with weight decay \(10^{-4}\), global gradient clipping at 1.0, learning rate \(2\times10^{-4}\), batch size 128, and bfloat16 mixed precision. Training runs use up to 50,000 optimizer steps, checkpoints every 1,000 steps, subset validation every 2,000 steps, full validation at epoch boundaries, a 12-epoch limit, and plateau stopping.

The actor-relevance label generator uses fixed weights \(w_d=1.0\), \(w_v=1.5\), \(w_{\mathrm{ttc}}=1.25\), \(w_{\ell}=0.75\), \(w_{\mathrm{vru}}=1.5\), and \(w_o=0.5\). The distance scale is \(20.0\,\mathrm{m}\), the route-corridor radius is \(3.5\,\mathrm{m}\), the TTC threshold is \(3.0\,\mathrm{s}\), and the minimum closing speed is \(0.1\,\mathrm{m/s}\). Ties are broken by ego-relative distance. Slot matching uses no distance cutoff; equivalently, $\tau_{\mathrm{match}}=\infty$. This keeps the matching rule deterministic and avoids introducing an additional tuned threshold, but it may assign distant predicted slots to sidecar actors early in training. The matched regression losses are therefore interpreted as part of the complete sidecar recipe rather than as an independently optimized detector-training objective.
\section{Results}
\label{sec:results}

We evaluate held-out waypoint prediction on the test split; lower ADE/FDE is better. All learned configurations are reported as mean $\pm$ sample standard deviation over three seeds. Non-deployable diagnostic configurations are marked with $^\dagger$ and use simulator-derived actor information at evaluation. Relative FDE reductions in Table \ref{tab:main_results} are computed against the same-seed plain waypoint-only RGB baseline unless a stronger control is explicitly named. Section \ref{sec:Route-Level-Paired-Analysis} also reports a conservative paired comparison against the matched no-teacher non-sidecar RGB control. 

\begin{table}[t]
\captionsetup{font=footnotesize,skip=2pt}
\centering
\caption{Main open-loop waypoint-prediction results on the exact test split. Values are mean $\pm$ sample standard deviation over three seeds. Relative FDE reduction is computed seed-wise relative to the same-seed plain waypoint-only RGB baseline. The waypoint-student/no-teacher configuration is the matched non-sidecar RGB control used for conservative comparisons in Sections \ref{sec:Route-Level-Paired-Analysis} and \ref{sec:Actor-Conditioned Slice Analysis}. }

\label{tab:main_results}

\scriptsize
\setlength{\tabcolsep}{2pt}
\renewcommand{\arraystretch}{1.08}
\setlength{\arrayrulewidth}{0.4pt}

\begin{tabularx}{\columnwidth}{|>{\raggedright\arraybackslash}X|c|c|c|}
\hline
Configuration
& \makecell[c]{ADE\\(m) $\downarrow$}
& \makecell[c]{FDE\\(m) $\downarrow$}
& \makecell[c]{Rel. FDE\\red. $\uparrow$} \\
\hline

Plain waypoint-only RGB
& $0.846{\pm}0.02$ & $1.815{\pm}0.02$ & -- \\
\hline
Forward/final weighted RGB
& $0.864{\pm}0.02$ & $1.848{\pm}0.02$ & $-1.81\%$ \\
\hline
Shuffled-teacher control
& $0.817{\pm}0.02$ & $1.772{\pm}0.02$ & $2.39\%$ \\
\hline
Waypoint student, no teacher (matched ctrl.)
& $0.803{\pm}0.02$ & $1.716{\pm}0.02$ & $5.48\%$ \\
\hline
Waypoint student, teacher-aligned
& $0.802{\pm}0.02$ & $1.716{\pm}0.02$ & $5.46\%$ \\
\hline
RU-sidecar, no teacher
& $0.608{\pm}0.02$ & $1.223{\pm}0.01$ & $32.6\%$ \\
\hline
RU-sidecar, no actor-future loss
& $0.921{\pm}0.02$ & $1.797{\pm}0.02$ & $0.99\%$ \\
\hline
RU-sidecar, no selector loss
& $0.89{\pm}0.02$ & $1.74{\pm}0.02$ & $4.13\%$ \\
\hline
RU-sidecar, teacher-aligned
& $0.594{\pm}0.02$ & $1.186{\pm}0.15$ & $34.7\%$ \\
\hline
Oracle sidecar-selected actor diagnostic$^\dagger$
& $0.561{\pm}0.02$ & $1.146{\pm}0.02$ & $36.84\%$ \\
\hline
Simulator-state actor-dynamics diagnostic$^\dagger$
& $0.103{\pm}0.02$ & $0.174{\pm}0.02$ & $90.42\%$ \\
\hline
\end{tabularx}
\end{table}

\subsection{Waypoint Prediction} \label{sec:results_main}

Table~\ref{tab:main_results} reports the main open-loop waypoint-prediction results. The waypoint-only RGB baseline obtains $1.815 \pm 0.02$ m test FDE. Reweighting the waypoint loss toward later horizons does not improve performance, indicating that the main gain is not explained by final-horizon loss weighting alone. The shuffled-teacher control gives only a modest improvement, so teacher-associated signals should not be credited automatically without a valid sample-level teacher relationship.

The full road-user sidecar recipe provides the main deployable improvement. Without teacher alignment, it reduces test FDE to 1.223 $\pm$ 0.01 m. This is a 32.6\% reduction relative to the plain waypoint-only RGB baseline and a 28.7\% reduction relative to the matched no-teacher non-sidecar RGB control. With teacher alignment, it obtains 1.186 $\pm$ 0.15 m FDE, corresponding to a 34.7\% reduction relative to the plain baseline and a 30.9\% reduction relative to the matched teacher-aligned non-sidecar control. Because the teacher-aligned result has substantially larger seed variability and only a small mean gain over the no-teacher RU-sidecar model, we treat teacher alignment as secondary. The main supported result is therefore the no-teacher RU-sidecar recipe.

The oracle sidecar-selected actor diagnostic$\dagger$ and simulator-state
actor-dynamics diagnostic$\dagger$ estimate the value of reliable road-user information. The oracle sidecar-selected actor diagnostic$\dagger$ obtains 0.561 $\pm$ 0.02 m test ADE and 1.146 $\pm$ 0.02 m test FDE. The simulator-state actor-dynamics diagnostic obtains 0.103 $\pm$ 0.02 m test ADE and 0.174 $\pm$ 0.02 m test FDE. These diagnostic configurations are not deployable camera-first systems. Their role is to show that accurate road-user state is highly predictive for the logged waypoint target and that camera-based models still recover and use this information only partially.

\begin{table*}[t]
\captionsetup{font=footnotesize,skip=2pt}
\centering
\small
\setlength{\tabcolsep}{5.0pt}
\renewcommand{\arraystretch}{1.1}
\setlength{\arrayrulewidth}{0.4pt}

\caption{Route-level paired improvement over the waypoint-only RGB baseline on 1,494 test routes. For each route and seed, $\Delta$FDE is computed as $\mathrm{FDE}_{RGB}-\mathrm{FDE}_{m}$; positive values indicate improvement. The reported route-level values average the paired improvements across the three seeds before computing win counts and bootstrap confidence intervals. Confidence intervals are 95\% route-bootstrap intervals for mean route-level $\Delta$FDE.}
\label{tab:route_paired}

\begin{tabular}{|l|c|c|c|c|c|}
\hline
\textbf{Model} &
\textbf{Mean \(\Delta\)FDE (m)} &
\textbf{Median \(\Delta\)FDE (m)} &
\textbf{95\% CI (m)} &
\textbf{FDE wins} &
\textbf{Win rate} \\
\hline

Forward/final-horizon weighted RGB
& $-0.034$ & $-0.021$ & $[-0.042,\,-0.025]$ & $641/1494$ & $42.90\%$ \\
\hline

Shuffled-teacher waypoint control
& $+0.033$ & $+0.078$ & $[+0.013,\,+0.053]$ & $897/1494$ & $60.04\%$ \\
\hline

No-teacher waypoint student
& $+0.132$ & $+0.148$ & $[+0.113,\,+0.149]$ & $1069/1494$ & $71.55\%$ \\
\hline

Teacher-aligned waypoint student
& $+0.112$ & $+0.142$ & $[+0.094,\,+0.130]$ & $1047/1494$ & $70.08\%$ \\
\hline

RU-sidecar, no teacher
& $+0.854$ & $+0.806$ & $[+0.823,\,+0.885]$ & $1445/1494$ & $96.72\%$ \\
\hline

RU-sidecar, teacher-aligned
& $+0.891$ & $+0.823$ & $[+0.861,\,+0.922]$ & $1448/1494$ & $96.92\%$ \\
\hline

Oracle sidecar-selected actor diagnostic\(\dagger\)
& $+0.642$ & $+0.611$ & $[+0.610,\,+0.674]$ & $1360/1494$ & $91.03\%$ \\
\hline

Simulator-state actor-dynamics diagnostic\(\dagger\)
& $+1.979$ & $+1.991$ & $[+1.937,\,+2.022]$ & $1494/1494$ & $100.00\%$ \\
\hline



\end{tabular}
\end{table*}

\subsection{Route-Level Paired Analysis} \label{sec:Route-Level-Paired-Analysis}

Table~\ref{tab:route_paired} reports route-level paired improvements over the same-seed waypoint-only RGB baseline. Positive $\Delta$FDE means that a configuration has lower route-averaged FDE than the baseline on the same route. This analysis checks whether the improvement is broad across routes rather than dominated by a small number of long or frequent routes.

The full no-teacher RU-sidecar model improves over the waypoint-only RGB baseline on 1445/1494 routes, corresponding to a 96.72\% route win rate. The full road-user teacher-aligned model improves on 1448/1494 routes, corresponding to a 96.92\% route win rate. These route-level results support the main conclusion that road-user supervision gives a broad improvement across held-out routes. The small difference between teacher-aligned and no-teacher route wins is treated as secondary because the teacher-aligned FDE has larger seed variability. The route-level averages are not expected to match the sample-weighted ADE/FDE ordering in Table~\ref{tab:main_results}, because each route receives equal weight here regardless of the number of valid samples on that route. The same weighting difference explains why the oracle sidecar-selected actor diagnostic$^\dagger$ can have lower sample-weighted FDE than RU-sidecar in Table \ref{tab:main_results} but a smaller route-level mean $\Delta$FDE in Table \ref{tab:route_paired}.

We also compute route-level paired improvements against the matched no-teacher non-sidecar RGB control, the waypoint student without teacher alignment. This comparison is more conservative than the plain waypoint-only baseline because this control already reduces sample-weighted test FDE from 1.815 m to 1.716 m. The RU-sidecar model without teacher alignment improves on 1417/1494 routes, with mean paired $\Delta$FDE of 0.722 m, median paired $\Delta$FDE of 0.565 m, and a 95\% route-bootstrap confidence interval of [0.686, 0.759] m. Thus, the route-level gain remains broad even under the stronger baseline and is not an artifact of choosing the weakest RGB baseline. The teacher-aligned RU-sidecar model gives similar route-level gains, improving on 1423/1494 routes relative to the teacher-aligned non-sidecar control.

\begin{figure}[t]
\centering
\captionsetup{font=footnotesize,skip=2pt}
\includegraphics[width=\columnwidth]{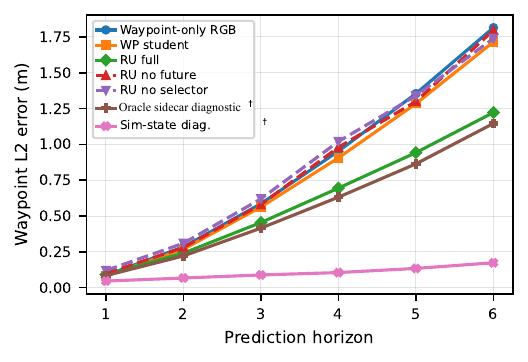}
\caption{Horizon-wise L2 waypoint error on the exact test split. The full road-user sidecar supervision (RU-sidecar) model reduces error across the prediction horizon relative to the waypoint-only RGB and waypoint-student baselines. Removing actor-future supervision or selector supervision largely removes this gain, with both ablation curves returning close to the waypoint-only RGB baseline.}


\label{fig:horizon_l2}
\end{figure}

\subsection{Horizon-Wise and Error-Component Behavior}
\label{sec:horizon_results}

Figure~\ref{fig:horizon_l2} shows that road-user sidecar supervision (RU-sidecar) reduces waypoint error across all six prediction horizons. Removing either actor-future supervision or selector supervision returns the curve close to the waypoint-only RGB baseline, consistent with Table~\ref{tab:main_results}. The non-deployable diagnostic curves remain lower, indicating that the tested RGB models do not close the privileged-to-camera gap.

\begin{table}[t]
\centering
\captionsetup{font=footnotesize,skip=2pt}
\caption{Actor-conditioned slice analysis on the exact test split. FDE values are sample-weighted means over three seeds. The matched no-teacher non-sidecar control is the waypoint student without teacher alignment. Reduction is computed as $100 \times (\mathrm{FDE}_{\mathrm{ctrl}}-\mathrm{FDE}_{\mathrm{RU}})/\mathrm{FDE}_{\mathrm{ctrl}}$. Actor-count rows use valid sidecar-actor counts from the exact-evaluation metadata. VRU denotes vulnerable road user.}
\label{tab:actor_slices}

\resizebox{\columnwidth}{!}{%
\begin{tabular}{|l|r|r|r|r|r|}
\hline
\textbf{Slice} & \textbf{Samples} & \textbf{Plain WP} & \textbf{Matched ctrl.} & \textbf{RU} & \textbf{Reduction} \\
\hline
\textbf{All test samples} & 467728 & 1.815 & 1.716 & 1.223 & 28.7\% \\
\hline
\textbf{1--3 valid sidecar actors} & 34758 & 1.027 & 1.056 & 0.830 & 21.4\% \\
\hline
\textbf{$\geq$4 valid sidecar actors} & 432970 & 1.878 & 1.769 & 1.254 & 29.1\% \\
\hline
\textbf{VRU present} & 73669 & 2.119 & 2.026 & 1.418 & 30.0\% \\
\hline
\textbf{High relevance score present} & 191641 & 1.161 & 1.121 & 0.980 & 12.6\% \\
\hline
\end{tabular}%
}
\end{table}

\subsection{Actor-Conditioned Slice Analysis}
\label{sec:Actor-Conditioned Slice Analysis}

Table~\ref{tab:actor_slices} reports actor-conditioned slice results on the exact test split. The slices are defined from sidecar metadata rather than model predictions, so all compared models are evaluated on the same sample sets. Because the available exact-evaluation metadata provides valid sidecar-actor counts rather than a verified image-visibility mask, we describe the actor-count rows as valid sidecar-actor slices. 

The RU-sidecar model improves over the matched no-teacher non-sidecar RGB control in every nonempty actor-conditioned slice. The reduction is 29.1\% for samples with at least four valid sidecar actors and 30.0\% when a vulnerable road user is present, compared with 21.4\% for samples with one to three valid sidecar actors. The high-relevance-score slice also improves, but by a smaller margin of 12.6\%. The high-relevance-score slice should not be interpreted as a generic hard-scene slice. It marks the presence of at least one actor with high target-conditioned sidecar relevance; its absolute FDE can be lower than the all-sample average because this slice may include slower or more structured interactions. We therefore interpret relative reduction within each slice rather than comparing
absolute FDE across slices. These results support the interpretation that RU-sidecar contributes to the gains in actor-rich and VRU-present scenes, while also showing that the improvement is not uniform across all actor-conditioned subsets.

\subsection{Learned Actor Diagnostics and Oracle Coverage}

Learned actor diagnostics are reported only for RU-sidecar models because no-sidecar controls receive no actor-sidecar supervision; their actor heads, when instantiated, are uncalibrated and are not used as learned actor diagnostics. The no-teacher RU-sidecar model obtains Recall@3/Recall@5 of 0.915/0.958 and Precision@3/Precision@5 of 0.304/0.225 against sidecar-defined relevance labels. The teacher-aligned RU-sidecar model obtains similar Recall@3/Recall@5 of 0.914/0.960 and Precision@3/Precision@5 of 0.303/0.223. Selected-actor future prediction remains poor, with Dyn-FDE of 9.529 m and 9.116 m for the no-teacher and teacher-aligned models, respectively. These actor diagnostics should not be read as evidence of reliable deployable multi-agent forecasting. The actor-future metric evaluates absolute future-state prediction after slot matching, whereas the waypoint planner uses predicted actor-future tokens as latent conditioning features. The high Dyn-FDE values show that the actor-future head is not a calibrated actor forecaster. However, Table \ref{tab:main_results} and Fig. \ref{fig:horizon_l2} show that the corresponding auxiliary loss is important for waypoint prediction: removing $L_dyn$ returns FDE close to the waypoint-only baseline. We therefore interpret the actor-future head as a representation-shaping and planner-conditioning mechanism, not as a deployable forecasting module.

Separately, the sidecar label generator’s top-5 selected set is designed for high coverage rather than high binary precision. On the test split, it covers 95.97\% of sidecar-relevant actors, with 16.72\% positive precision. These statistics characterize the sidecar critical-set construction, not learned RGB actor-selection performance and not an upper bound on learned selector precision.

\section{Conclusion}

We studied whether simulator-derived RU-sidecar improves camera-first open-loop waypoint prediction while preserving RGB-only inference. Under a matched student-interface architecture and route-disjoint evaluation protocol, the RU-sidecar model without teacher alignment reduces FDE to 1.223 $\pm$ 0.01 m, improving by 32.6\% over the plain waypoint-only RGB baseline and by 28.7\% over the matched no-teacher non-sidecar RGB control. Route-level paired analysis shows that the RU-sidecar model remains broadly better under this stronger comparison, improving on 1417/1494 held-out routes with mean paired $\Delta$FDE of 0.722 m. Actor-conditioned slice analysis shows improvements across all nonempty actor-conditioned slices, with especially large reductions in scenes with at least four valid sidecar actors and in VRU-present scenes. Teacher alignment gives a slightly lower mean FDE but is less stable across seeds, so the primary supported claim is the no-teacher sidecar supervision recipe. Non-deployable diagnostics show that exact road-user state remains much more predictive than camera-derived actor representations. The study is limited to open-loop simulation diagnostics and does not establish closed-loop safety, crash avoidance, real-world transfer, or deployment readiness.

\bibliographystyle{IEEEtran}
\bibliography{references}

@inproceedings{10.1109/ICRA.2018.8460487,
author = {Codevilla, Felipe and Miiller, Matthias and L\'{o}pez, Antonio and Koltun, Vladlen and Dosovitskiy, Alexey},
title = {End-to-End Driving Via Conditional Imitation Learning},
year = {2018},
publisher = {IEEE Press},
url = {https://doi.org/10.1109/ICRA.2018.8460487},
doi = {10.1109/ICRA.2018.8460487},
booktitle = {2018 IEEE International Conference on Robotics and Automation (ICRA)},
pages = {1–9},
numpages = {9},
location = {Brisbane, Australia}
}

@InProceedings{pmlr-v100-chen20a,
  title = 	 {Learning by Cheating},
  author =       {Chen, Dian and Zhou, Brady and Koltun, Vladlen and Kr\"ahenb\"uhl, Philipp},
  booktitle = 	 {Proceedings of the Conference on Robot Learning},
  pages = 	 {66--75},
  year = 	 {2020},
  editor = 	 {Kaelbling, Leslie Pack and Kragic, Danica and Sugiura, Komei},
  volume = 	 {100},
  series = 	 {Proceedings of Machine Learning Research},
  month = 	 {30 Oct--01 Nov},
  publisher =    {PMLR},
  url = 	 {https://proceedings.mlr.press/v100/chen20a.html}
}

@INPROCEEDINGS{9711506,
  author={Zhang, Zhejun and Liniger, Alexander and Dai, Dengxin and Yu, Fisher and Van Gool, Luc},
  booktitle={2021 IEEE/CVF International Conference on Computer Vision (ICCV)}, 
  title={End-to-End Urban Driving by Imitating a Reinforcement Learning Coach}, 
  year={2021},
  volume={},
  number={},
  pages={15202-15212},
  doi={10.1109/ICCV48922.2021.01494}}

@ARTICLE{9863660,
  author={Chitta, Kashyap and Prakash, Aditya and Jaeger, Bernhard and Yu, Zehao and Renz, Katrin and Geiger, Andreas},
  journal={IEEE Transactions on Pattern Analysis and Machine Intelligence}, 
  title={TransFuser: Imitation With Transformer-Based Sensor Fusion for Autonomous Driving}, 
  year={2023},
  volume={45},
  number={11},
  pages={12878-12895},
  doi={10.1109/TPAMI.2022.3200245}}

@inproceedings{10.5555/3600270.3600713,
author = {Wu, Penghao and Jia, Xiaosong and Chen, Li and Yan, Junchi and Li, Hongyang and Qiao, Yu},
title = {Trajectory-guided control prediction for end-to-end autonomous driving: a simple yet strong baseline},
year = {2022},
isbn = {9781713871088},
publisher = {Curran Associates Inc.},
address = {Red Hook, NY, USA},
booktitle = {Proceedings of the 36th International Conference on Neural Information Processing Systems},
articleno = {443},
numpages = {14},
location = {New Orleans, LA, USA},
series = {NIPS '22}
}

@inproceedings{chen2022lav,
  title={Learning from all vehicles},
  author={Chen, Dian and Kr{\"a}henb{\"u}hl, Philipp},
  booktitle={CVPR},
  year={2022}
}

@inproceedings{Renz2022CORL,
    author       = {Katrin Renz and Kashyap Chitta and Otniel-Bogdan Mercea and A. Sophia Koepke and Zeynep Akata and Andreas Geiger},
    title        = {PlanT: Explainable Planning Transformers via Object-Level Representations},
    booktitle    = {Conference on Robotic Learning (CoRL)},
    year         = {2022}
}

@misc{shao2023lmdrive,
      title={LMDrive: Closed-Loop End-to-End Driving with Large Language Models}, 
      author={Hao Shao and Yuxuan Hu and Letian Wang and Steven L. Waslander and Yu Liu and Hongsheng Li},
      year={2023},
      eprint={2312.07488},
      archivePrefix={arXiv},
      primaryClass={cs.CV}
}

@inproceedings{mile2022,
  title     = {Model-Based Imitation Learning for Urban Driving},
  author    = {Anthony Hu and Gianluca Corrado and Nicolas Griffiths and Zak Murez and Corina Gurau
   and Hudson Yeo and Alex Kendall and Roberto Cipolla and Jamie Shotton},
  booktitle = {Advances in Neural Information Processing Systems ({NeurIPS})},
  year = {2022}
}

\end{document}